\documentclass[journal,twoside]{IEEEtran}
\usepackage{graphicx}
\usepackage{multirow}
\usepackage{textcomp}
\usepackage{subcaption}
\usepackage{amsmath}
\usepackage{breqn}
\usepackage[usestackEOL]{stackengine}
\usepackage{mathtools}
\usepackage{lipsum}
\usepackage{pdflscape}
\usepackage{tabularx,booktabs, longtable}
\usepackage{hyperref}
\usepackage{float}
\usepackage{makecell}

\begin{document}

\title{Automated Detection and Forecasting of COVID-19 using Deep Learning Techniques: A Review  }

\author{Afshin~Shoeibi,
        Marjane~Khodatars,
        Mahboobeh~Jafari,
        Navid~Ghassemi,
        Delaram~Sadeghi,
        Parisa~Moridian,
        Ali~Khadem,
        Roohallah~Alizadehsani,~\IEEEmembership{Member,~IEEE,}
        Sadiq~Hussain,
        Assef~Zare,
        Zahra~Alizadeh~Sani,
        Fahime~Khozeimeh,
        Saeid~Nahavandi,~\IEEEmembership{Fellow,~IEEE,},
        U.~Rajendra~Acharya, and 
        Juan~M.~Gorriz
\thanks{A. Shoeibi, M. Khodatars, M. Jafari, D. Sadeghi, and P. Moridian are with the Data Science and Computational Intelligence Institute, University of Granada, Spain.
(Corresponding author: Afshin Shoeibi, email: afshin.shoeibi@gmail.com).}
\thanks{N. Ghassemi is with the Computer Engineering Department, Ferdowsi University of Mashhad, Mashhad, Iran.}
\thanks{R. Alizadehsan, F. Khozeimeh, and S. Nahavandi. are with the Institute for Intelligent Systems Research and Innovation (IISRI), Deakin University, Victoria 3217, Australia.}
\thanks{A. Khadem is with the Faculty of Electrical Engineering, K. N. Toosi University of Technology, Tehran, Iran.}
\thanks{Sadiq Hussain is System Administrator at Dibrugarh University, Assam, India, 786004.}
\thanks{A. Zare is with Faculty of Electrical Engineering, Gonabad Branch, Islamic Azad University, Gonabad, Iran.}
\thanks{Z. Alizadeh Sani is with Rajaie Cardiovascular Medical and Research Center, and Iran University of Medical Sciences, Tehran, Iran.}
\thanks{U. R. Acharya is with the School of Mathematics, Physics and Computing, University of Southern Queensland, Springfield, Australia.}
\thanks{J. M. Gorriz is with the Data Science and Computational Intelligence Institute, University of Granada, Spain, and Dept. of Psychiatry. University of Cambridge, UK.}
}

\markboth{}{Shoeibi \MakeLowercase{\textit{et al.}}: Automated Detection and Forecasting of COVID-19 using Deep Learning Techniques}

\maketitle

\begin{abstract}
Coronavirus, or COVID-19, is a hazardous disease that has endangered the health of many people around the world by directly affecting the lungs. COVID-19 is a medium-sized, coated virus with a single-stranded RNA, and also has one of the largest RNA genomes and is approximately 120 nm. The X-Ray and computed tomography (CT) imaging modalities are widely used to obtain a fast and accurate medical diagnosis. Identifying COVID-19 from these medical images is extremely challenging as it is time-consuming and prone to human errors. Hence, artificial intelligence (AI) methodologies can be used to obtain consistent high performance. Among the AI methods, deep learning (DL) networks have gained popularity recently compared to conventional machine learning (ML). Unlike ML, all stages of feature extraction, feature selection, and classification are accomplished automatically in DL models. In this paper, a complete survey of studies on the application of DL techniques for COVID-19 diagnostic and segmentation of lungs is discussed, concentrating on works that used X-Ray and CT images. Additionally, a review of papers on the forecasting of coronavirus prevalence in different parts of the world with DL is presented. Lastly, the challenges faced in the detection of COVID-19 using DL techniques and directions for future research are discussed.
\end{abstract}


\IEEEpeerreviewmaketitle

\section{Introduction}

The novel COVID-19 virus came to light in December 2019 in Wuhan Province, China, where it originated from animals and quickly spread around the world \cite{a1}. In January 2020, world health organization (WHO) announced the epidemic of COVID-19 as a threat to public health. In March 2020, it announced the Corona pandemic \cite{cydivid}. Coronaviruses include various types that mainly occur in animals. A type of Coronavirus, called SARS-CoV-2, is transmitted from bats to humans, threatening human health throughout the world \cite{cydivid}. SARS-CoV2 might stay alive on different surfaces from a few hours to several days. Clinical studies show that the incubation period of this virus is 1-14 days \cite{cydivid}. Recently, a new type of Coronavirus called Delta with a short incubation period has involved many people, and it has more dangerous complications \cite{delta}. 

\begin{figure}[t]
    \centering
    \includegraphics[width=3.1in ]{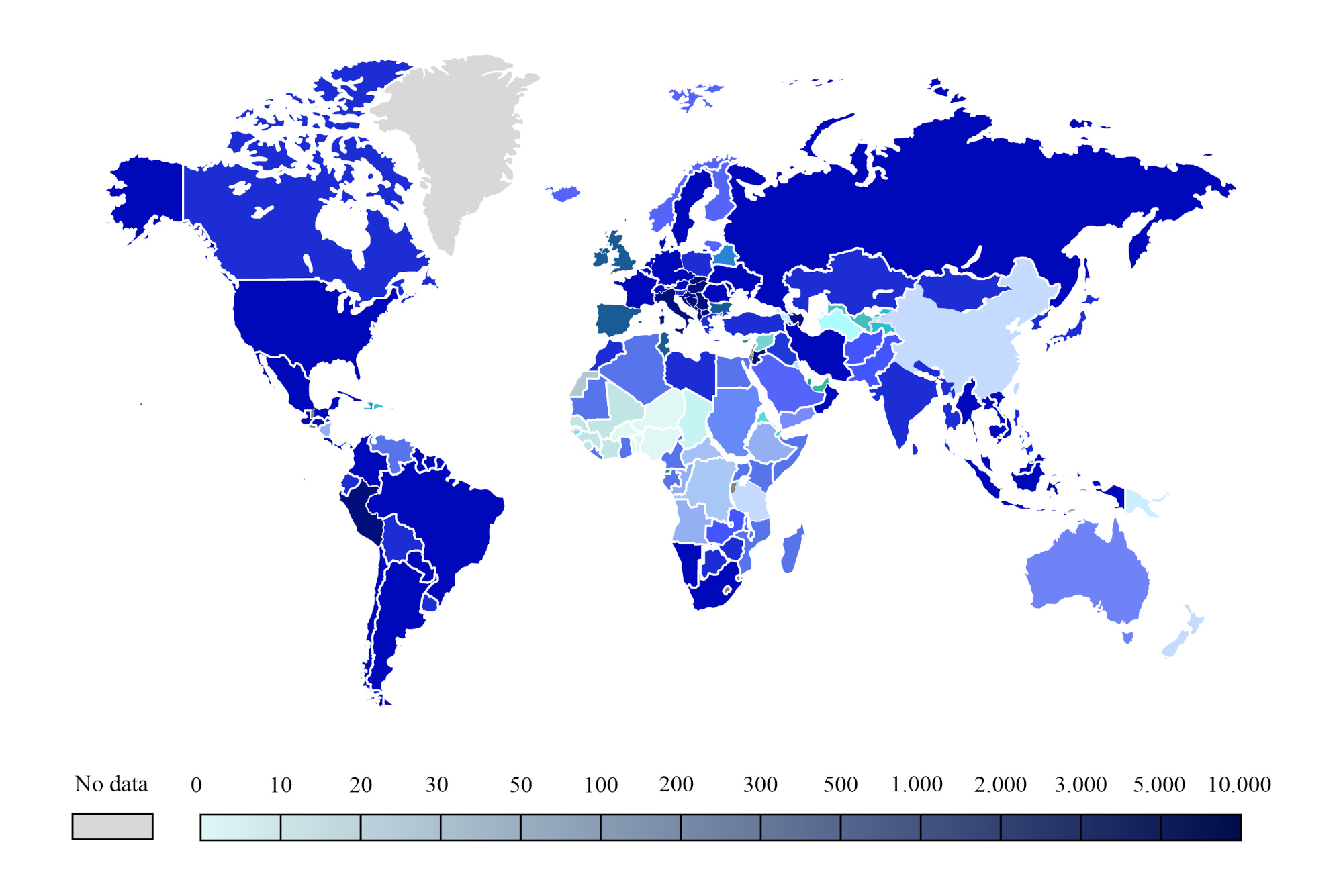}  
    
    \caption{The latest detailed statistics of COVID-19 infected people worldwide \cite{a7}.}
    \label{fig:one}
\end{figure}

The easiest way to transmit SARS-CoV-2 is through the air and physical contact, such as hand contact with an infected person \cite{a2}. The virus inserts itself into the lung cells through the respiratory system and replicates there, destroying these cells \cite{a3}. COVID-19 comprises an ribonucleic acid (RNA) and is very difficult to diagnose and treat due to its mutation characteristics \cite{a4}. The most common symptoms of SARS-CoV-2 include fever, cough, and shortness of breath, dizziness, headache, and muscle aches \cite{cydivid}. The virus is so perilous and can provoke the death of people with weakened immune systems \cite{a6}. Infectious disease specialists and physicians around the world are working to discover a treatment for the disease.  COVID-19 is currently the leading cause of death for thousands of countries worldwide, including the USA, Spain, Italy, China, the United Kingdom, Iran, and others. Figure \ref{fig:one} shows the latest number of infected people worldwide due to COVID-19.


Currently, various methods have been proposed for fast diagnosis of the Coronavirus. Among the proposed methods, WHO has introduced the real-time reverse transcription polymerase chain reaction (PT-PCR) test as the gold standard of early diagnosis of COVID-19 \cite{a8}. Also, imaging methods like X-Ray, CT, and ultrasound are of significance for COVID-19 diagnosis.

According to the WHO, all diagnoses of corona disease must be confirmed by RT-PCR \cite{a7}. However, testing with RT-PCR is highly time-consuming, and this issue is risky for people with COVID-19. Hence, first, medical imaging is carried out for the primary detection of COVID-19, then the RT-PCR test is performed to aid the physicians in making final accurate detection. Two medical imaging techniques, X-ray, CT-scan, and ultrasound are employed to diagnose COVID-19 \cite{an74,an89}.

X-ray modality is the first procedure to diagnose COVID-19, which has the advantage of being inexpensive and low-risk from radiation hazards to human health \cite{a11}. In the X-ray method, detecting COVID-19 is a relatively complicated task. In these images, the radiologist must attentively recognize the white spots that contain water and pus, which is very prolonged and problematic. A radiologist or specialist doctor may also mistakenly diagnose other diseases, such as pulmonary tuberculosis, as COVID-19 \cite{a12}.

The X-ray procedure has a high error rate. CT modality has a higher contrast compared to X-Ray \cite{an89}. CT data of the patients suffering from SARS-CoV-2 demonstrate pulmonary parenchyma destruction, including interstitial inflammation and extensive consolidation accurately \cite{an99}. To detect the Coronavirus, numerous CT slices are recorded from each patient, which their analysis is challenging. To this end, the specialists try to remove the slices that do not contain important information.

Lung ultrasound is another Coronavirus diagnosis method \cite{an116}. Ultrasound plays an important role in diagnosing and treating the Coronavirus because it is not a radiative method. Also, ultrasound can assess different organs and systems, including the heart, arteries, and kidneys, that might have been damaged due to COVID-19 \cite{an116}. 

In recent years, applications of AI in medicine have led to a variety of studies aiming to diagnose varied diseases, including brain tumors from magnetic resonance (MR) images \cite{a16,a17}, multiple types of brain disorders such from electroencephalography (EEG) \cite{a18}, breast cancer from mammographic images \cite{a20,a21} and pulmonary diseases such as Covid-19 from X-Ray \cite{an74} and CT \cite{an89}. In the last decade, DL, a branch of ML, has changed the expectations in many applications of AI in data processing by reaching human-level accuracies \cite{humanlevel} in many tasks, including medical image analysis \cite{humanlevelmri}.

In this paper, an overview of COVID-19 diagnostic approaches utilizing DL networks is presented. Section II explains the search strategy, and various DL models developed for COVID-19 detection are described in Section III. Section IV of the DL techniques used for the detection, segmentation, and prediction of COVID-19 patients. Section V discusses the reviewed papers on diagnosis, segmentation, and prediction of COVID-19 patients. Challenges in diagnosing, segmentation, and prediction of COVID-19 patients are provided in Section VI. Finally, the summary and future work are delineated in Section VII.
    
\section{Search Strategy}
\begin{figure}[t]
    \centering
    \includegraphics[width=1.65in ]{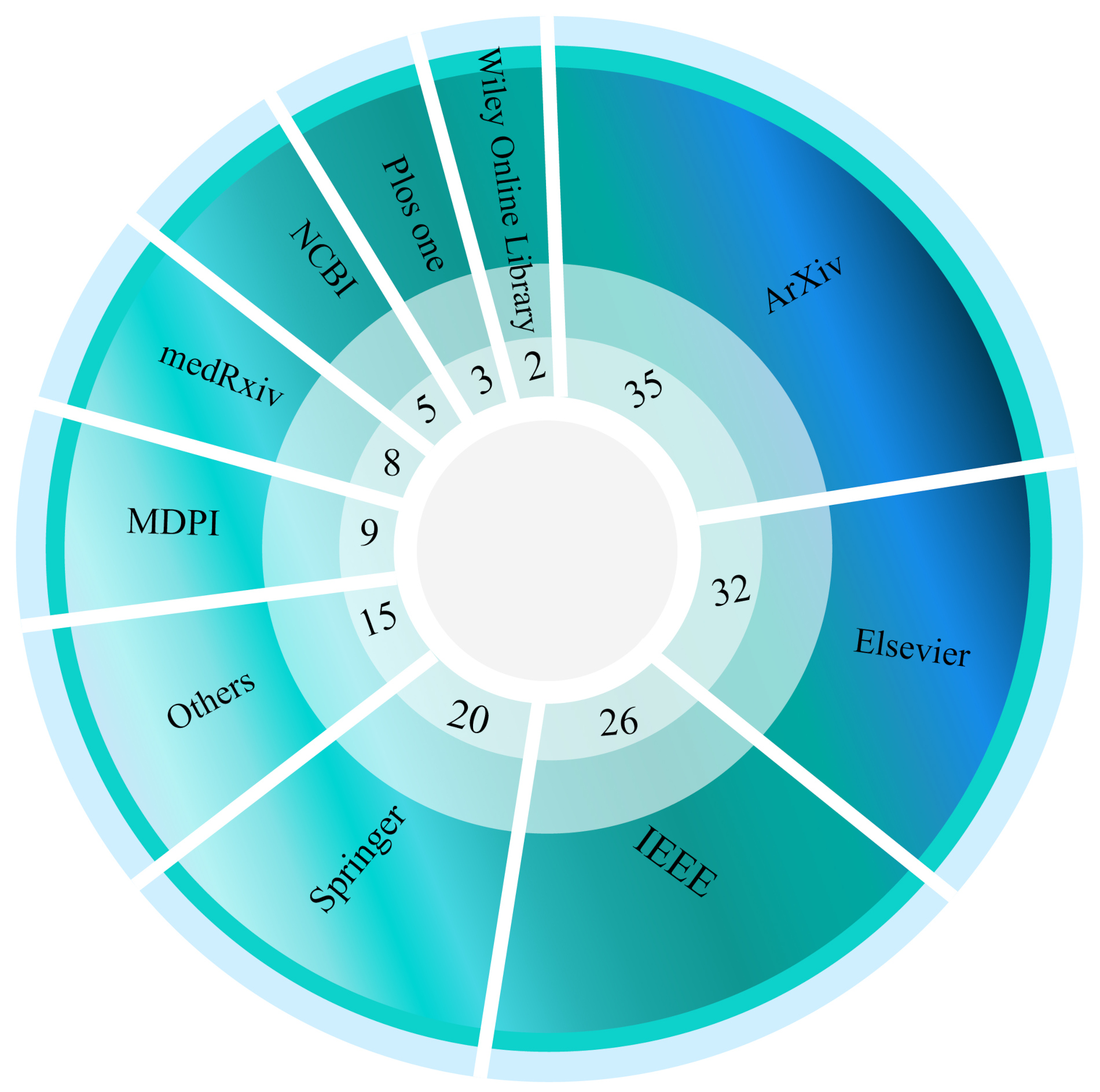}
    
    \caption{Number of papers published on COVID-19 using DL techniques.}
    \label{fig:two}
\end{figure}

In this study, valid databases, including IEEE Xplore, ScienceDirect, SpringerLink, ACM, and ArXiv, have been used to search for Covid-19 papers. Moreover, a more detailed Google Scholar search is employed. The articles are selected using the keywords “COVID-19”, “Corona Virus”, “Deep Learning”, “Segmentation”, “Forecasting”, “Attention Deep Learning”, “Transformer Deep  Learning”, “Data Fusion”, and “Graph Deep Learning”. The latest selection of papers is done with the mentioned keywords on September 19th, 2021. Figure \ref{fig:two} indicates the number of papers published or indexed by COVID-19 using DL techniques using various databases.

\section{Deep Learning Techniques for COVID-19 Detection}
Conventional machine learning and DL are the two main branches of AI, but DL is essentially a more advanced version of conventional ML. Various DL network architectures have been extensively used in research papers to diagnose the COVID-19 accurately using publicly available databases.
\begin{figure}[t]
    \centering
    \includegraphics[width=3.5in ]{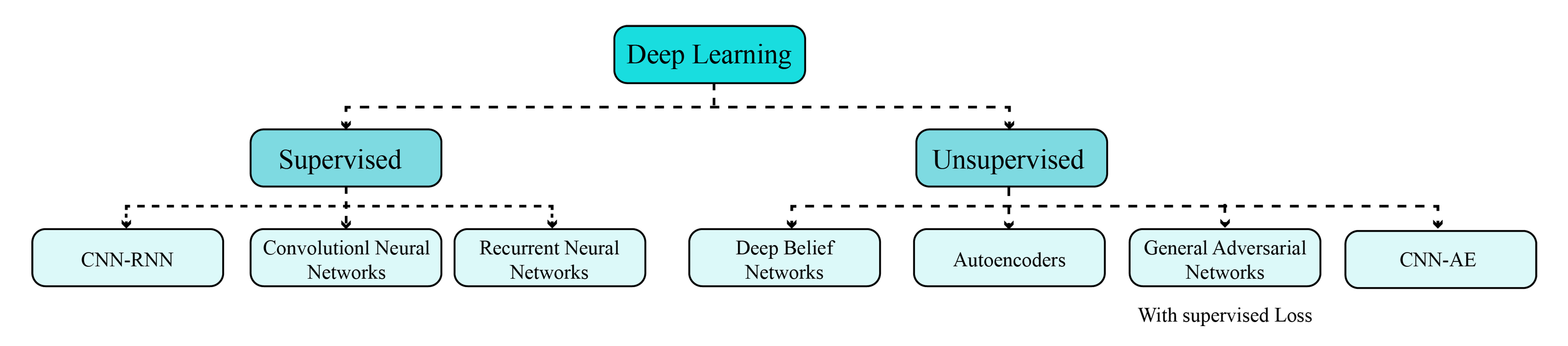}
    
    \caption{Illustration of various DL methods used for COVID-19 detection.}
    \label{fig:three}
\end{figure}
Many of well-known DL architectures, such as convolutional neural networks (CNNs), recurrent neural networks (RNNs), Autoencoders (AEs), deep belief networks (DBNs), generative adversarial networks (GANs), and also some hybrid networks such as CNN-RNN and CNN-AE have been developed for automated detection of COVID-19. Figure \ref{fig:three} shows the subcategories of DL networks.

\section{Computer Aided Diagnosis System (CADS) for COVID-19 Detection}
\begin{figure}[t]
    \centering
    \includegraphics[width=3.3in ]{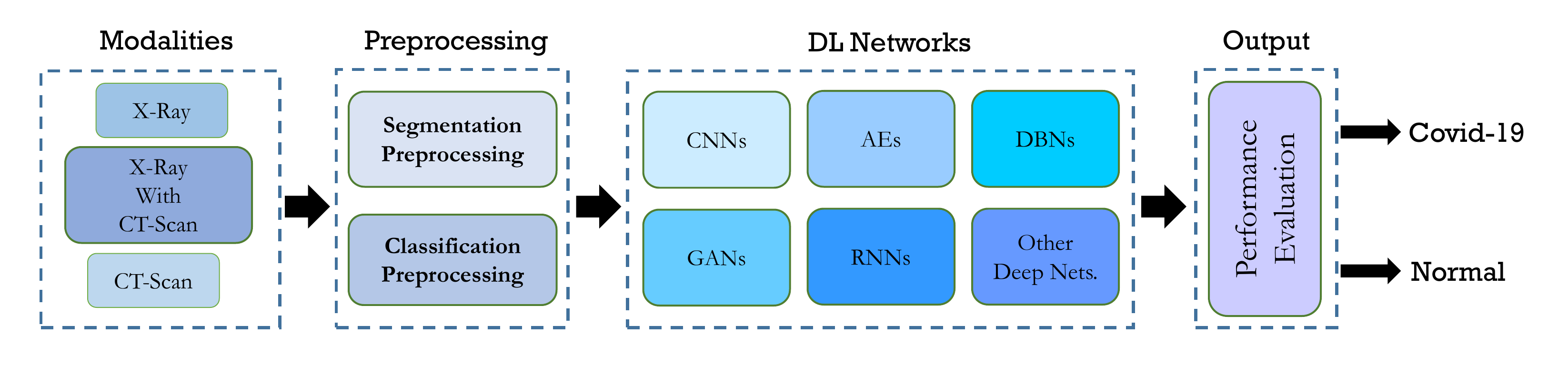}
    
    \caption{Block diagram for COVID-19 detection using DL technique.}
    \label{fig:four}
\end{figure}

In prior research papers, many CADS have been developed applying DL methods on X-ray and CT images; these systems can be categorized by their application into two categories: (i) classification and (ii) segmentation. In classification-based CADS, the main objective is to identify COVID-19 patients, which involves the process of extracting and selecting the most informative features and classifying using DL. However, in the second type, an image of an infected person is given to the system for segmentation of an area of interest. Manual segmentation of medical images takes considerable time; thus, applying machine learning models is crucially paramount. Among the most important segmentation models, the several types of fuzzy clustering methods \cite{a24,a25} and DL ones such as U-Net \cite{unet} can be denoted. In the CADS, with the segmentation approach, patients' CT-Scan images and their manual segments labeled by doctors are fed to the DL network. Then, during the training process, the DL network is trained on manual segments to segment raw input images. The components of DL-based CADS for COVID-19 detection are shown in Figure \ref{fig:four}. In the following section, we will first mention the important data available for COVID-19; then, the DL methods used in the review research are introduced.

\subsection{Public Databases used for COVID-19 Detection and Forecasting}
Several public databases (X-ray and CT images) are available for the detection and segmentation of COVID-19, some of them are listed in Table \ref{tabledataone}. Also, the datasets related to predicting the COVID-19 spread in leading countries of the world are shown in Table \ref{tabledatatwo}.

\begin{table}[t]
\caption{Public databases used for COVID-19 detection.}
\label{tabledataone}
\centering
\resizebox{3.5in}{!}{
\begin{tabular}{|c|c|c|}
\hline
Dataset & Modality & Link\\
\hline
J. P. Cohen’s GitHub \cite{t1p1} &X-ray and CT	&https://github.com/ieee8023/covid-chestxray-dataset\\
\hline
European Society of Radiology	&X-ray and CT	&https://www.eurorad.org/advanced-search?search=COVID\\
\hline
SIRM	&X-ray and CT	&https://www.sirm.org/category/senza-categoria/covid-19\\
\hline
BSTI	&X-ray and CT	&https://www.bsti.org.uk/covid-19-resources\\
\hline
UCSD-AI4H \cite{t1p5} &CT	&https://github.com/UCSD-AI4H/COVID-CT\\
\hline
MedSeg	&CT	&http://medicalsegmentation.com/covid19\\
\hline
Kaggle	&X-ray and CT	&https://www.kaggle.com/datasets?search=covid\\
\hline
\multicolumn{1}{|c|}{\multirow{2}{*}{\shortstack[1]{Point-of-Care Ultrasound\\(POCUS) \cite{t1p8}}}} &
\multicolumn{1}{c|}{\multirow{2}{*}{\shortstack[1]{Lung Ultrasound\\Images and Videos}}} &
\multicolumn{1}{c|}{\multirow{2}{*}{\shortstack[1]{https://github.com/jannisborn/covid19\_pocus\_ultrasound}}}\\
\multicolumn{1}{|c|}{} & \multicolumn{1}{c|}{} & \multicolumn{1}{c|}{}\\
\hline
\multicolumn{1}{|c|}{\multirow{2}{*}{\shortstack[1]{Actualmed COVID-19 Chest\\X-ray Dataset Initiative}}} & \multicolumn{1}{c|}{\multirow{2}{*}{\shortstack[1]{X-ray}}} & \multicolumn{1}{c|}{\multirow{2}{*}{\shortstack[1]{https://github.com/agchung/Actualmed-COVID-chestxray-dataset}}}\\
\multicolumn{1}{|c|}{} & \multicolumn{1}{c|}{} & \multicolumn{1}{c|}{}\\
\hline
\multicolumn{1}{|c|}{\multirow{2}{*}{\shortstack[1]{COVID-19 Chest X-ray\\Dataset Initiative}}} & \multicolumn{1}{c|}{\multirow{2}{*}{\shortstack[1]{X-ray}}} & \multicolumn{1}{c|}{\multirow{2}{*}{\shortstack[1]{https://github.com/agchung/Figure1-COVID-chestxray-dataset}}}\\
\multicolumn{1}{|c|}{} & \multicolumn{1}{c|}{} & \multicolumn{1}{c|}{}\\
\hline
\multicolumn{1}{|c|}{\multirow{2}{*}{\shortstack[1]{Georgia State University’s\\Panacea Lab \cite{t1p11}}}} & \multicolumn{1}{c|}{\multirow{2}{*}{\shortstack[1]{Twitter Chatter\\Dataset}}} & \multicolumn{1}{c|}{\multirow{2}{*}{\shortstack[1]{https://github.com/thepanacealab/covid19\_twitter}}}\\
\multicolumn{1}{|c|}{} & \multicolumn{1}{c|}{} & \multicolumn{1}{c|}{}\\
\hline
Twitter COVID‐19 CXR dataset	&X-ray	&https://twitter.com/ChestImaging\\
\hline
COVID-19 \cite{t1p13} &CT	&https://github.com/KevinHuRunWen/COVID-19\\
\hline
COVIDx	\cite{t1p14}&X-ray	&https://github.com/lindawangg/COVID-Net\\
\hline
\end{tabular}}
\end{table}

\begin{table}[t]
\caption{Public COVID-19 forecasting databases used for forecasting.}
\label{tabledatatwo}
\centering
\resizebox{3.5in}{!}{
\begin{tabular}{|c|c|c|}
\hline
Dataset	&Modality	&Link\\
\hline
China CDC Weekly	&Daily Number of Cases in China	&http://weekly.chinacdc.cn/news/TrackingtheEpidemic.htm\\
\hline
\multicolumn{1}{|c|}{\multirow{2}{*}{\shortstack[1]{The Ministry of Health and Family\\Welfare (Government of India)}}} & \multicolumn{1}{c|}{\multirow{2}{*}{\shortstack[1]{Daily Number of Cases in India}}} & \multicolumn{1}{c|}{\multirow{2}{*}{\shortstack[1]{ https://www.mohfw.gov.in}}}\\
\multicolumn{1}{|c|}{} & \multicolumn{1}{c|}{} & \multicolumn{1}{c|}{}\\
\hline
Johns Hopkins University	&Tracking COVID-19 Spread	&https://systems.jhu.edu\\
\hline
WHO COVID-19 Dashboard	&Global Statistics	&https://covid19.who.int\\
\hline
\multicolumn{1}{|c|}{\multirow{2}{*}{\shortstack[1]{U.S. CDC}}} & \multicolumn{1}{c|}{\multirow{2}{*}{\shortstack[1]{Daily Number of Cases in U.S.}}} & \multicolumn{1}{c|}{\multirow{2}{*}{\shortstack[1]{https://www.cdc.gov/coronavirus/2019-ncov/cases-updates/cases-in-us.html\\https://www.cdc.gov/coronavirus/2019-ncov/covid-data/data-visualization.htm}}}\\
\multicolumn{1}{|c|}{} & \multicolumn{1}{c|}{} & \multicolumn{1}{c|}{}\\
\hline
Worldometer	&Global Collection	&https://www.worldometers.info/coronavirus\\
\hline

Open Source COVID-19	&Global Collection	&http://open-source-covid-19.weileizeng.com\\
\hline
Painel Coronavírus	&Daily Number of Cases in Brazil	&https://covid.saude.gov.br\\
\hline
GOV.UK	&Daily Number of Cases in UK	&https://coronavirus.data.gov.uk\\
\hline
\multicolumn{1}{|c|}{\multirow{2}{*}{\shortstack[1]{Ministero della Salute}}} & \multicolumn{1}{c|}{\multirow{2}{*}{\shortstack[1]{Daily Number of Cases in Italy}}} & \multicolumn{1}{c|}{\multirow{2}{*}{\shortstack[1]{http://www.salute.gov.it/portale/nuovocoronavirus/\\homeNuovoCoronavirus.jsp?lingua=english}}}\\
\multicolumn{1}{|c|}{} & \multicolumn{1}{c|}{} & \multicolumn{1}{c|}{}\\
\hline
\multicolumn{1}{|c|}{\multirow{3}{*}{\shortstack[1]{Ministry of health}}} & \multicolumn{1}{c|}{\multirow{3}{*}{\shortstack[1]{Daily Number of Cases in Spain}}} & \multicolumn{1}{c|}{\multirow{3}{*}{\shortstack[1]{https://www.mscbs.gob.es/profesionales/saludPublica/ccayes/alertasActual/\\nCov-China/situacionActual.htm\\https://cnecovid.isciii.es/covid19}}}\\
\multicolumn{1}{|c|}{} & \multicolumn{1}{c|}{} & \multicolumn{1}{c|}{}\\
\multicolumn{1}{|c|}{} & \multicolumn{1}{c|}{} & \multicolumn{1}{c|}{}\\
\hline

\end{tabular}}
\end{table}

\subsection{Deep Learning Methods}

This section is devoted to describing the methods applied in papers briefly. First, well-known network structures for both classification and segmentation are discussed; then, models which are applied for forecasting are explained and lastly, new trends and state-of-the-art methods are presented.

\subsubsection{Classification Models}

\textsc{CNNs and Pre-trained models}

The primary issue in training the deep models is the concern of overfitting that occurs from the gap between the limited number of training samples and a large number of learnable parameters. Convolutional networks try to overcome this by using convolutional layers \cite{cnn}. CNNs require minimal pre-processing by considering the 2-dimensional (2D) images as input, and hence it is designed to retain and utilize the structural information among neighboring pixels or voxels. A differentiable function is utilized to transform one volume of actions by each layer to the other as it is a sequence of layers structurally. 

\begin{figure}[t]
    \centering
    \includegraphics[width=3.1in ]{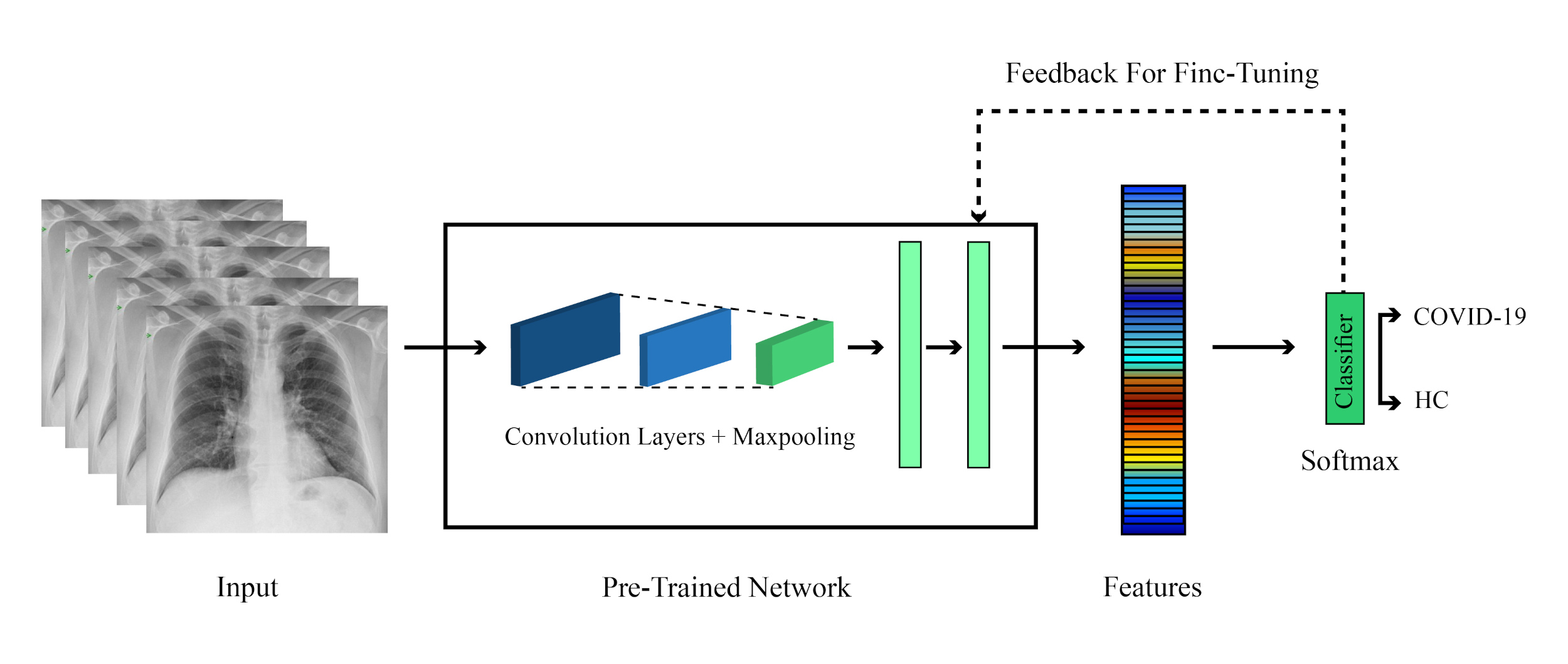}
    
    \caption{Overall diagram of pre-trained methods.}
    \label{fig:six}
\end{figure}

While convolutional layers work as some sort of workaround for the issues of deep neural networks, training CNNs properly still needs a massive amount of data, and also designing their structure is itself a time-consuming process. To overcome these problems, researchers usually use a pre-trained version of well-known network architecture. Figure \ref{fig:six} shows how the pre-trained networks are used; also, some of the most used network architectures are : AlexNet \cite{a53}, Visual Geometry Group (VGG) network \cite{vgg}, GoogLeNet \cite{a102}, ResNet \cite{resnet}, DenseNet \cite{densenet}, and, SqueezeNet \cite{a74}.


\subsubsection{Generative Adversarial Networks (GAN)}

A primary problem in training deep models is limits in dataset size. Using generative models for data augmentation is one solution to this issue. Due to the high quality of generated data, GANs have attracted attention in the medical imaging community \cite{ganmed}. The basic idea in training a GAN is a simple minimax game, in which one network tries to distinguish between real data and generates one, and the other tries to create data undistinguishable by the first network \cite{a44}, therefor creating images similar to real data.

\subsubsection{Segmentation Models}

A wide variety of  DL models have been developed for the segmentation of the lung region to detect COVID-19 in patients accurately. Among these models, FCN network \cite{fcn}, SegNet \cite{a149}, U-Net \cite{unet}, and Res2Net \cite{a139} DL models are widely used for the segmentation of lungs. In this section, some of these models are briefly discussed.

\textsc{SegNet}

Generally, in segmentation techniques, a network created for classification is chosen, and the FC layers of that network are removed; the resulting network is called the encoder network. Then a decoder is created to transform these low-resolution maps to the original resolution. In SegNet \cite{a149}, the decoder is created such that for each down-sampling layer in the encoding section, an up-sampling layer is positioned in the decoder. These layers, unlike the deconvolution layers of FCN networks, are not capable of learning, and the values are placed at the locations from which the corresponding max-pooling layer is extracted, and the rest of the output cells become zero. 

\textsc{U-Net}

\begin{figure}[t]
    \centering
    \includegraphics[width=3.1in ]{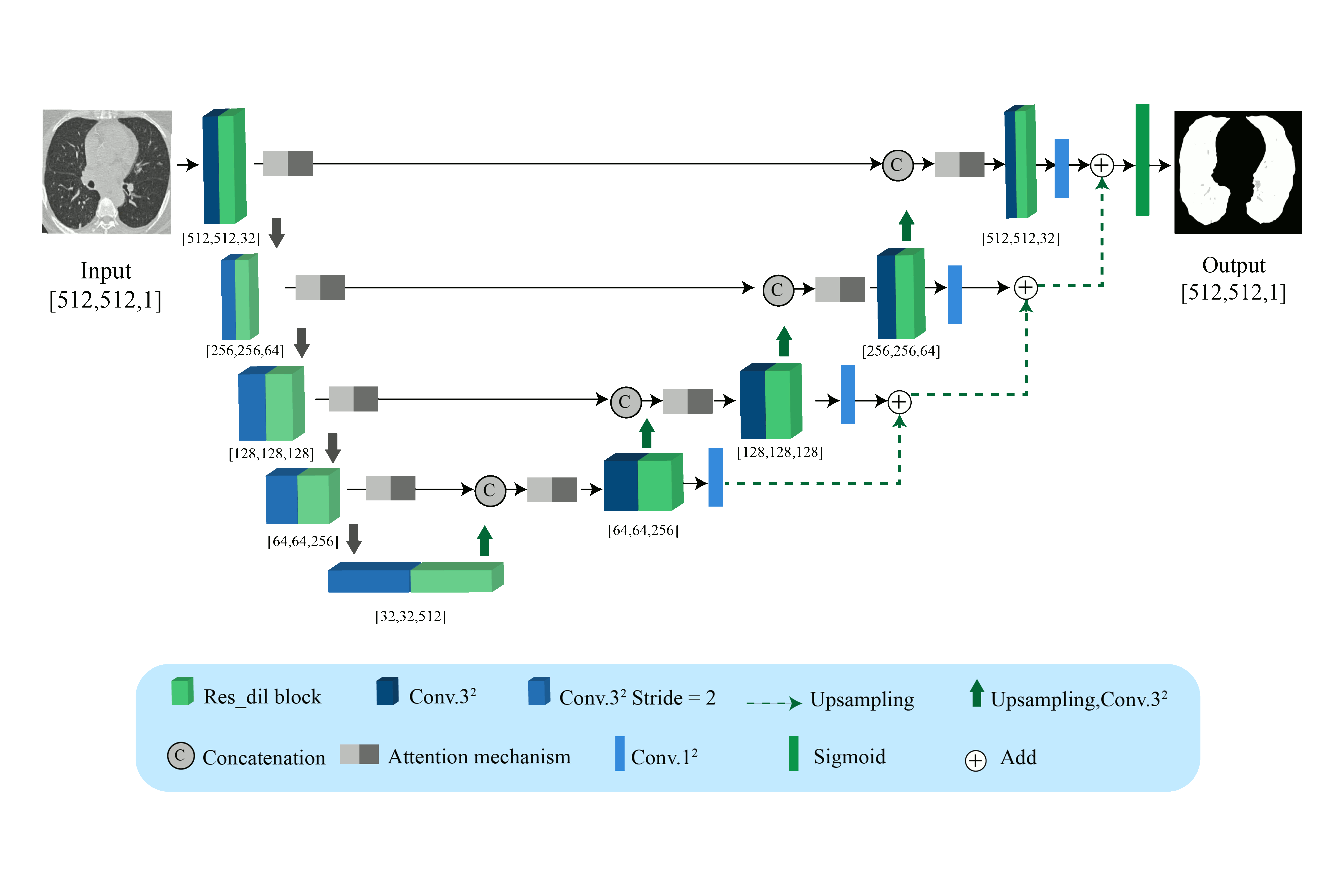}
    
    \caption{A typical U-Net architecture used to segment the lung in COVID-19 patients \cite{unetshape}.}
    \label{fig:unet}
\end{figure}

The U-Net network \cite{unet}, like SegNet, consists of the identical numbers of pooling and up-sampling layers, but the network utilizes trainable deconvolution layers. Also, in this network, there is a corresponding skip connection between the up-sampling and down-sampling layers. Figure \ref{fig:unet} shows a general form of U-Net architecture used to segment the lung in COVID-19 patients.

\subsubsection{Forecasting Models}

\textsc{Recurrent Neural Network(RNN)}

A feed-forward neural network is extended to create RNN, aiming to capture the long term dependencies and features from the sequential and time-series data. The most commonly used RNN is the long-short term memory (LSTM), which composed of a memory cell, a forget cell, the input gate, and output gate. These gates make the decision that which information needs to be remembered or discarded from the memory cell and also organizes the activation signals from different sources.

LSTM decides whether to keep or remove the memory by using these gates; also, unlike vanilla RNN, LSTM can preserve the potential long term dependencies. One LSTM variant is Gated Recurrent Unit (GRU) \cite{a48}, which integrates the forget and input gates into a single update gate and combines the memory cell state and the hidden state into one state. Update gate makes a decision on the amount of information to be added or discarded, and the reset gate decides on how much earlier information is to be forgotten. This technique makes GRU simpler than LSTM. 

    



\subsubsection{Advanced AI methods for Diagnosis of COVID-19}

\textsc{Deep Attention Learning}

The DL methods based on attention mechanisms have attracted attention recently \cite{attenp}. The models based on attention mechanism are concentrated on a subset of inputs (focus on certain parts of the input) that contain information regarding the tasks \cite{attenp}. Recently, attention models have been used in various applications, including classification, segmentation, and diagnosis of COVID-19. In \cite{attenp}, a DL model based on attention with the attention module of VGG-16 has been used to classify COVID-19 using X-Ray images. 

\textsc{Deep Transformer Learning}

Transformer models are another type of DL method, and some of their techniques include spatial transformer networks, graph transformer networks, recurrent spatial transformer networks, and Polar transformer networks. In \cite{tranp}, COVID-19 has been diagnosed using ultrasound data based on the vision transformer (ViT) model. 

\textsc{Deep Fusion Techniques}

With the emergence of DL models, it has been tried to combine data fusion techniques and DL networks with medical objectives. In \cite{fusionp}, CT data have been extracted using four CNNs to classify Corona data. Then, feature integration and ranking techniques have been used to obtain effective features. Finally, the SVM classifier has been used.

\textsc{Graph Deep Learning}

The graph models based on DL are another class of new DL techniques that have been recently used in detecting the coronavirus. In \cite{graphp}, first, a 3D-CNN has been used to extract features from CT images. Then, a COVID-19 graph in GCN has been designed based on the features. Finally, these three DL models are combined to detect the COVID-19.

\section{Discussion}
\begin{figure}[t]
    \centering
    \includegraphics[width=1.65in ]{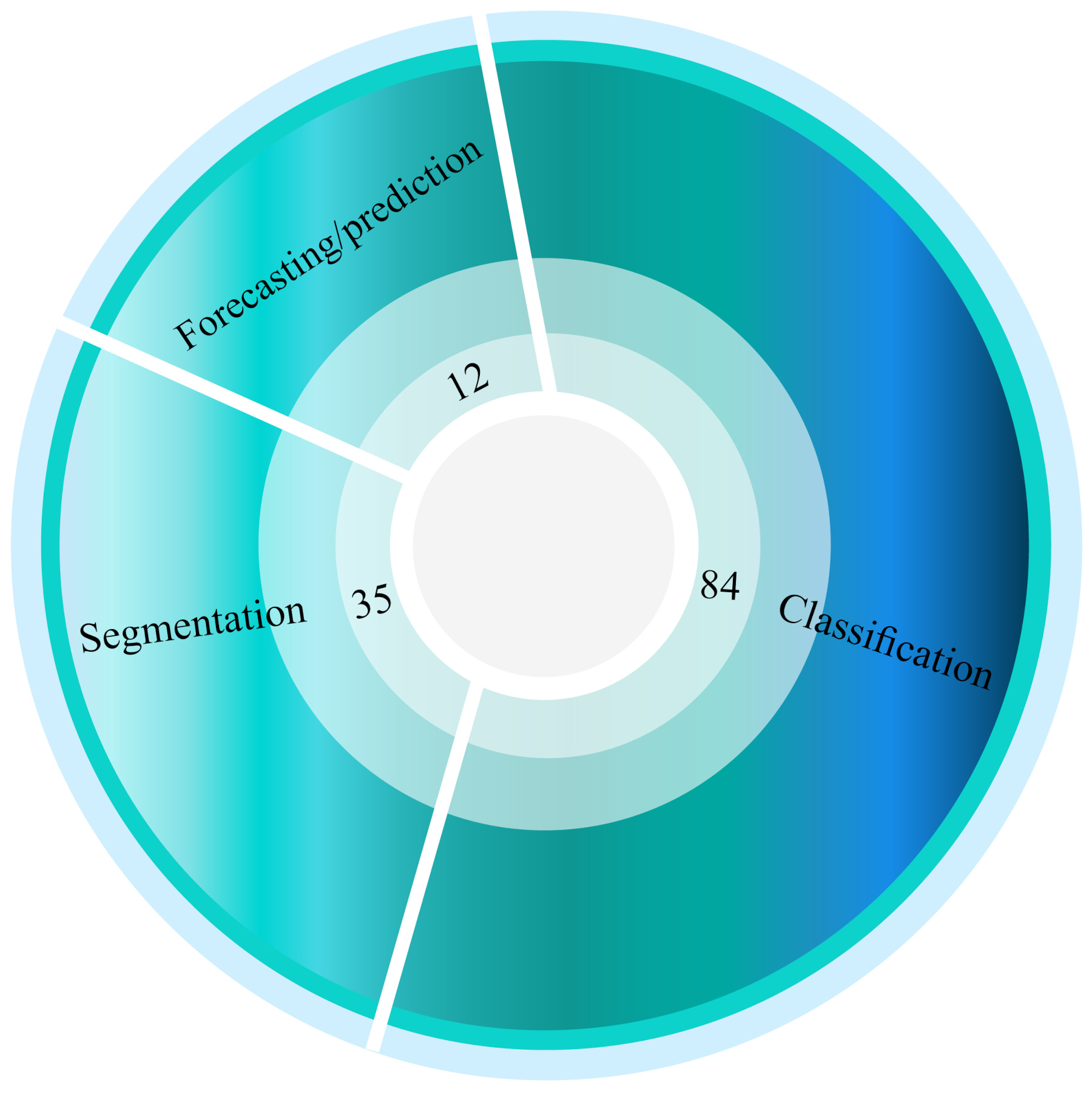}
    
    \caption{Total number of investigations conducted in the field of classification, segmentation, and prediction of COVID-19 patients using DL techniques.}
    \label{fig:21}
\end{figure}

The main focus of this work is to review the research papers that have worked on DL models for detection, segmentation of the lungs and also forecasting the spread of the COVID-19. The summary of works done on classification, segmentation, and forecasting are presented in Tables \ref{tableone}, \ref{tabletwo}, and \ref{tablethree}, respectively. Figure \ref{fig:21} depicts the total number of investigations conducted in the field of classification, segmentation, and forecasting of COVID-19 using DL models. It can be noted from the figure that most works have been done on the detection of COVID-19 patients, and the least works are done on forecasting due to the shortage of available public databases.

\onecolumn
\begin{landscape}

\tiny
\begin{longtable}{|c|c|c|c|c|c|c|c|c|c|c|c|}
    \caption{Summary of state-of-art DL techniques used for the automated detection of COVID-19 patients.}    
        \label{tableone}\\
      \hline
      \multicolumn{1}{|c|}{\multirow{2}{*}{\shortstack[1]{Work}}} & \multicolumn{1}{c|}{\multirow{2}{*}{\shortstack[1]{Dataset}}} & \multicolumn{1}{c|}{\multirow{2}{*}{\shortstack[1]{Modalities}}} & \multicolumn{1}{c|}{\multirow{2}{*}{\shortstack[1]{Number of Cases}}} & \multicolumn{1}{c|}{\multirow{2}{*}{\shortstack[1]{Preprocessing}}} & \multicolumn{1}{c|}{\multirow{2}{*}{\shortstack[1]{DNN toolbox}}} & \multicolumn{1}{c|}{\multirow{2}{*}{\shortstack[1]{DNN}}} & \multicolumn{1}{c|}{\multirow{2}{*}{\shortstack[1]{Number of\\Layers}}} & \multicolumn{1}{c|}{\multirow{2}{*}{\shortstack[1]{Classifier}}} & \multicolumn{1}{c|}{\multirow{2}{*}{\shortstack[1]{Post Processing}}} & \multicolumn{1}{c|}{\multirow{2}{*}{\shortstack[1]{K-Fold}}} & \multicolumn{1}{c|}{\multirow{2}{*}{\shortstack[1]{Performance \\Criteria (\%)}}}\\
    \multicolumn{1}{|c|}{} & \multicolumn{1}{c|}{} & \multicolumn{1}{c|}{} & \multicolumn{1}{c|}{} & \multicolumn{1}{c|}{} & \multicolumn{1}{c|}{} & \multicolumn{1}{c|}{} & \multicolumn{1}{c|}{} & \multicolumn{1}{c|}{} & \multicolumn{1}{c|}{} & \multicolumn{1}{c|}{} & \multicolumn{1}{c|}{}\\
\hline
    
     \end{longtable}
     \begin{longtable}{|c|c|c|c|c|c|c|c|c|c|c|c|}
    \caption{Summary of DL segmentation methods used for COVID-19 detection.}    
        \label{tabletwo}\\
      \hline
      \multicolumn{1}{|c|}{\multirow{2}{*}{\shortstack[1]{Work}}} & \multicolumn{1}{c|}{\multirow{2}{*}{\shortstack[1]{Dataset}}} & \multicolumn{1}{c|}{\multirow{2}{*}{\shortstack[1]{Modalities}}} & \multicolumn{1}{c|}{\multirow{2}{*}{\shortstack[1]{Number of Cases}}} & \multicolumn{1}{c|}{\multirow{2}{*}{\shortstack[1]{Preprocessing}}} & \multicolumn{1}{c|}{\multirow{2}{*}{\shortstack[1]{DNN toolbox}}} & \multicolumn{1}{c|}{\multirow{2}{*}{\shortstack[1]{DNN}}} & \multicolumn{1}{c|}{\multirow{2}{*}{\shortstack[1]{Number of\\Layers}}} & \multicolumn{1}{c|}{\multirow{2}{*}{\shortstack[1]{Classifier}}} & \multicolumn{1}{c|}{\multirow{2}{*}{\shortstack[1]{Post Processing}}} & \multicolumn{1}{c|}{\multirow{2}{*}{\shortstack[1]{K-Fold}}} & \multicolumn{1}{c|}{\multirow{2}{*}{\shortstack[1]{Performance \\Criteria (\%)}}}\\
    \multicolumn{1}{|c|}{} & \multicolumn{1}{c|}{} & \multicolumn{1}{c|}{} & \multicolumn{1}{c|}{} & \multicolumn{1}{c|}{} & \multicolumn{1}{c|}{} & \multicolumn{1}{c|}{} & \multicolumn{1}{c|}{} & \multicolumn{1}{c|}{} & \multicolumn{1}{c|}{} & \multicolumn{1}{c|}{} & \multicolumn{1}{c|}{}\\
\hline
    
     \end{longtable}
\end{landscape}
\twocolumn

\begin{table*}[t]
\caption{Summary of DL models developed for the forecasting spread of COVID-19.}
\label{tablethree}
\centering
\resizebox{7in}{!}{
\begin{tabular}{|c|c|c|c|c|c|c|c|c|}
\hline
\multicolumn{1}{|c|}{\multirow{2}{*}{\shortstack[1]{Work}}} & \multicolumn{1}{c|}{\multirow{2}{*}{\shortstack[1]{Dataset}}} & \multicolumn{1}{c|}{\multirow{2}{*}{\shortstack[1]{Preprocessing}}} & \multicolumn{1}{c|}{\multirow{2}{*}{\shortstack[1]{DNN\\Toolbox}}} & \multicolumn{1}{c|}{\multirow{2}{*}{\shortstack[1]{DNN}}} & \multicolumn{1}{c|}{\multirow{2}{*}{\shortstack[1]{Number\\of layers}}} & \multicolumn{1}{c|}{\multirow{2}{*}{\shortstack[1]{K-Fold}}} & \multicolumn{1}{c|}{\multirow{2}{*}{\shortstack[1]{Post\\Processing}}} & \multicolumn{1}{c|}{\multirow{2}{*}{\shortstack[1]{Performance\\Criteria}}}\\
\multicolumn{1}{|c|}{} & \multicolumn{1}{c|}{} & \multicolumn{1}{c|}{} & \multicolumn{1}{c|}{} & \multicolumn{1}{c|}{} & \multicolumn{1}{c|}{} & \multicolumn{1}{c|}{} & \multicolumn{1}{c|}{} & \multicolumn{1}{c|}{}\\
\hline
\multicolumn{1}{|c|}{\multirow{2}{*}{\shortstack[1]{\cite{a161}}}} & \multicolumn{1}{c|}{\multirow{2}{*}{\shortstack[1]{Johns Hopkins University and Canadian Health Authority}}} & \multicolumn{1}{c|}{\multirow{2}{*}{\shortstack[1]{WT, ADF test}}} & \multicolumn{1}{c|}{\multirow{2}{*}{\shortstack[1]{NA}}}  & \multicolumn{1}{c|}{\multirow{2}{*}{\shortstack[1]{LSTM}}} & \multicolumn{1}{c|}{\multirow{2}{*}{\shortstack[1]{3}}} & \multicolumn{1}{c|}{\multirow{2}{*}{\shortstack[1]{NA}}} & \multicolumn{1}{c|}{\multirow{2}{*}{\shortstack[1]{NA}}} & \multicolumn{1}{c|}{\multirow{2}{*}{\shortstack[1]{RMSE=45.70\\Acc=92.67\%}}}\\
\multicolumn{1}{|c|}{} & \multicolumn{1}{c|}{} & \multicolumn{1}{c|}{} & \multicolumn{1}{c|}{} & \multicolumn{1}{c|}{} & \multicolumn{1}{c|}{} & \multicolumn{1}{c|}{} & \multicolumn{1}{c|}{} & \multicolumn{1}{c|}{}\\
\hline
\end{tabular}}
\end{table*}

\begin{figure}[t]
    \centering
    \includegraphics[width=1.65in ]{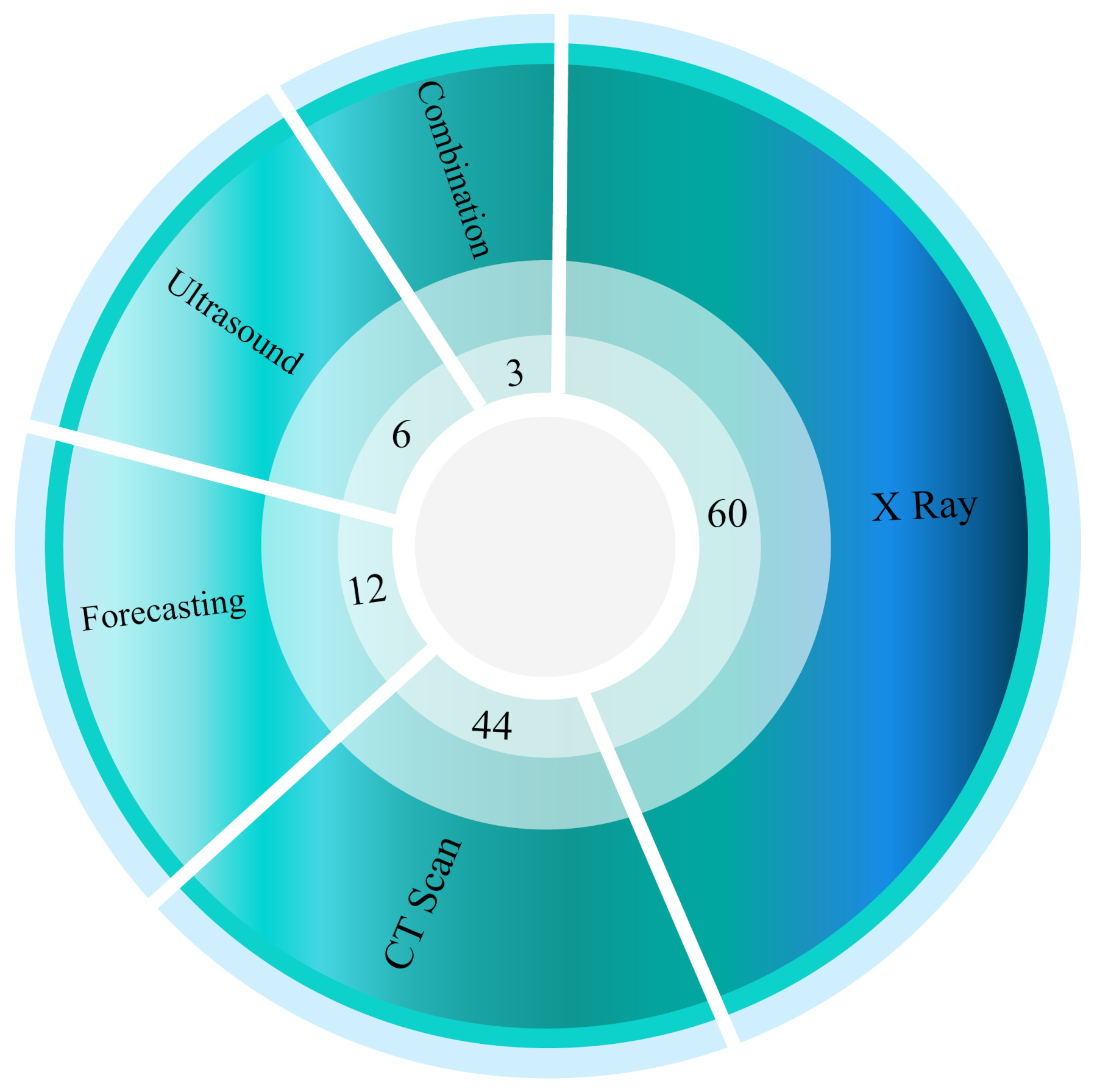}
    
    \caption{Number of datasets used for COVID-19 detection and prediction based on the published papers using DL methods. }
    \label{fig:22}
\end{figure}

The X-ray, ultrasound, and CT modalities have been used to develop DL models. Figure \ref{fig:22} shows the total number of times each modality is used in reviewed studies. It can be observed that most of the researchers have used X-ray images. This may be due to cheaper registration fees, and the fact that slice selection is not needed. Also, very few research papers have used combined modalities of X-ray and CT images due to the absence of such a comprehensive database.

\begin{figure}[t]
    \centering
    \includegraphics[width=1.65in ]{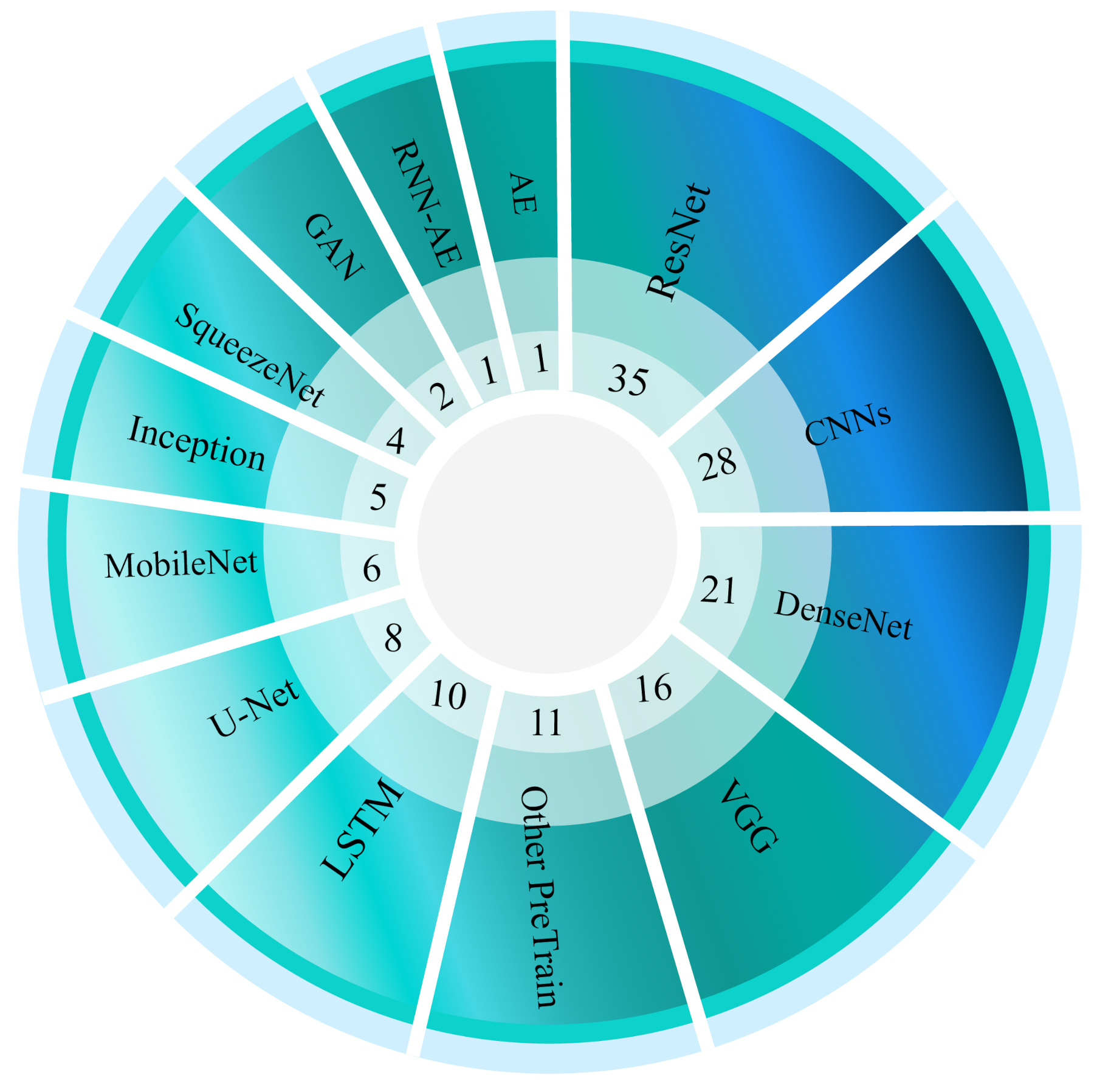}
    
    \caption{Number of DL architectures used for COVID-19 detection and prediction based on published papers.}
    \label{fig:23}
\end{figure}

Various DL models developed for the automated detection of COVID-19 patients are shown in Figure \ref{fig:23}. It can be noted from the figure that; different types of convolutional networks have been commonly used. Also, for the automated segmentation of lungs, various types of U-Net are more common.

\begin{figure}[t]
    \centering
    \includegraphics[width=1.65in ]{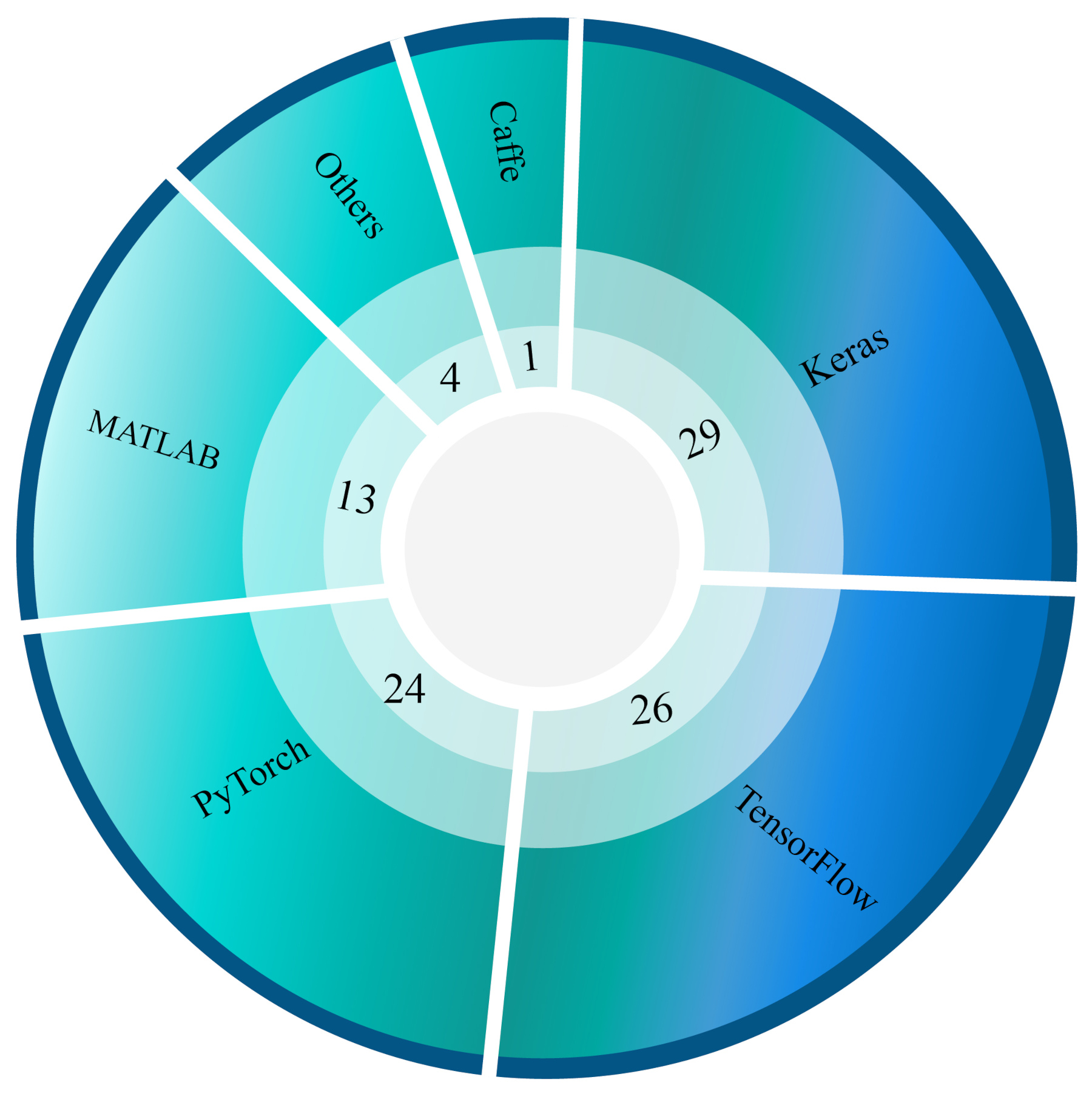}
    
    \caption{Number of DL tools used for COVID-19 detection and prediction based on the published papers. }
    \label{fig:24}
\end{figure}

Nowadays, a variety of toolboxes have been used to implement DL models. The number of toolboxes used for automated detection of COVID-19 by researchers is shown in Figure \ref{fig:24}. It can be noted that the Keras toolbox is the most widely used in reviewed papers; this can be due to its simplicity and also the availability of pre-trained models in this library, which are also used frequently by researchers.

\begin{figure}[t]
    \centering
    \includegraphics[width=1.65in ]{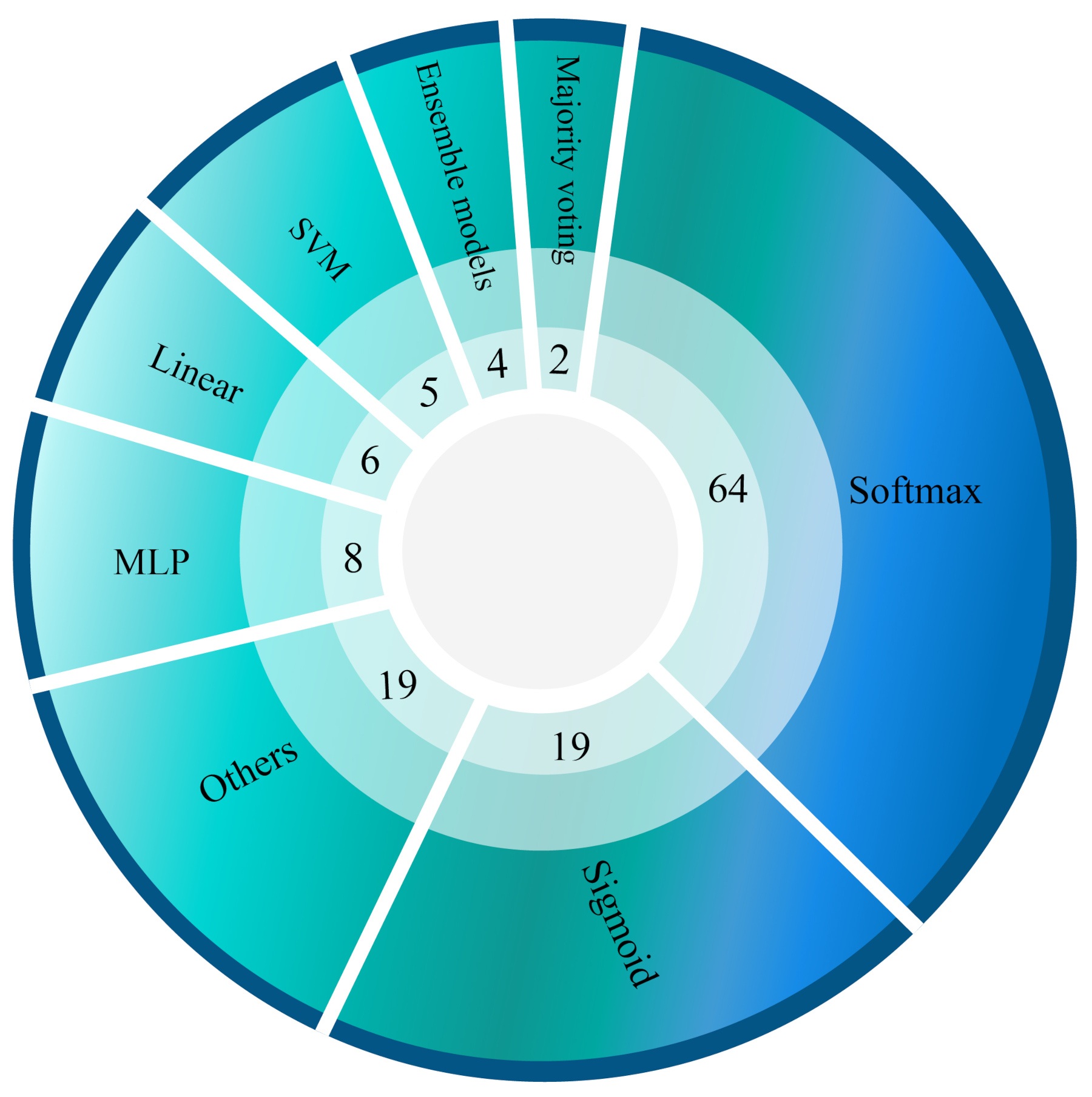}
    
    \caption{Illustration of number of various classifier algorithms used in DL networks for automated detection of COVID-19 patients.}
    \label{fig:25}
\end{figure}

The last part of this study is devoted to the classification algorithms developed using DL architectures. The softmax is most employed for automated detection of COVID-19 patients (Tables \ref{tableone} to \ref{tablethree}). Figure \ref{fig:25} shows the number of various classification algorithms used for automated detection of COVID-19 patients using DL techniques. 

\section{Challenges}

With the rapid growth and spread of COVID-19 globally, researchers have confronted many serious challenges in designing and implementing CADS to diagnose and also forecast the spread of the disease. The most significant challenges associated with COVID-19 are data availability, DL networks architecture fixing, and hardware resources. Lack of availability of a huge public database comprising X-ray and CT images is the first challenge. Due to the limited number of patient data, many researchers have used pre-trained networks such as GoogLeNet and AlexNet. Nevertheless, the number of studies conducted on forecasting is limited as it also requires a vast database.

One of the problems of employing pre-trained networks is that these models are often trained on the ImageNet database, which is entirely different from medical images. Hence, implementing efficient CADS to accurately and swiftly diagnose COVID-19 from X-Ray or CT images is still a challenging work. Physicians are not just convinced with X-ray or CT-scan images of patients to accurately diagnose COVID-19; they may use both modalities simultaneously. However, complete and comprehensive databases of the X-ray and CT-scan hybrid modalities for CADS research and implementation have not been provided for researchers in the machine learning scope. For this reason, researchers combine different X-ray and CT-Scan datasets from various datasets, which may disrupt network training. Yielding the combined X-Ray and CT-Scan datasets pave the path to help quickly identify COVID-19 alongside DL networks. The third challenge in the data section is the non-reporting of phenotypic information, such as age and gender. The utilization of this information can amend and enhance the performance of DL algorithms.

Table \ref{tabletwo} summarizes the DL-based segmentation algorithms aimed at identifying areas suspected of COVID-19 in the X-ray and CT-scan images. One of the obstacles with databases is the absence of manual or ground truths for COVID-19 image segmentation areas. Therefore, many researchers have delineated these areas with the help of radiologists and trained the models such as U-Net, which is time-consuming. Consequently, the presence of dedicated databases of segmented images will help to get the best-performing model. Also, it becomes easy to compare the performances with other authors who have worked on the same images. 

In order to predict the prevalence of corona using DL methods, the nature of the COVID-19 is still relatively unknown, and the probability of mutation is a big issue. Therefore, to predict the prevalence of the disease, many factors like the average age of the society, policies to impede the spread of the disease by countries, climatic conditions, and infection of neighbor/friend/family member.
 
Lack of access to appropriate hardware resources is another challenge. Implementing DL architectures in CADS for corona diagnosis demands strong hardware resources, which unfortunately is not ordinarily accessible for many researchers. Although tools such as Google Colab have partially obviated this problem, employing these tools in real medical applications is still challenging. For this reason, in most studies, researchers have not provided practical CADS systems such as web or Windows software to detect COVID-19.

\section{Conclusion and Future Works}

COVID-19 is an emerging pandemic disease that, in a short period of time, can severely endanger the health of many people throughout the world. It directly affects the lung cells, and if not accurately diagnosed early, can cause irreversible damage, including death. The disease is accurately detected by the specialists using X-ray, CT and ultrasound images together with RT-PCR results. Specialized physicians use RT-PCR as the gold standard for COVID-19 diagnosis based on WHO guidelines. The accurate diagnosis of COVID-19 is made using medical imaging methods, including X-Ray, CT, and Ultrasound beside RT-PCR. COVID-19 diagnosis by medical imaging always has some challenges for physicians. The high number of medical images associated with each patient, doctor's fatigue, and low contrast of the images are among some problems challenging the COVID-19 accurate diagnosis. The accurate diagnosis of COVID-19 has high importance, and lack of fast and accurate diagnosis can cause severe damages such as death.

To address this problem, researchers are working on some advanced DL techniques to diagnose COVID-19 accurately in the shortest possible time. In this study, a comprehensive review of the accomplished studies of COVID-19 diagnosis was carried out using DL networks. Frist, the public databases available to detection and prediction of COVID-19 are presented. In the following, the state-of-art DL techniques employed for the diagnosis, segmentation, and forecasting of the spread of COVID-19 are presented in Tables \ref{tableone}, \ref{tabletwo}, \ref{tablethree}, respectively.

In another section of the paper, some advanced DL methods such as attentions \cite{attenp}, transformers \cite{tranp}, fusion \cite{fusionp}, and graph \cite{graphp} have been introduced. In this section, an introduction to the method has been presented first, then COVID-19 diagnosis papers based on these methods are introduced. In the following, important information of each paper, including various types of DL research for COVID-19 (classifications, segmentation, and predictions), have been explored and compared. In the following, the number of modalities used in COVID-19 diagnosis, DL models, DL toolboxes, and classification algorithms are introduced.

The most critical challenge of COVID-19 diagnosis using DL techniques is mentioned in section 4. As mentioned, the most critical challenges of COVID-19 diagnosis are dataset, software, and hardware. One of the challenges to develop a robust and accurate COVID 19 diagnosis system is the availability of an extensive public database. We strongly feel that, with more public databases, better DL models can be developed by researchers to detection and prediction of the COVID19 accurately.

Recently, using fusion techniques such as feature fusion has had many improvements in medical applications \cite{fusion}. In future works, it is possible to use deep feature fusion techniques besides medical imaging modalities to diagnose COVID-19. In addition to that, it is possible to use some advanced DL methods such as zero-shot learning in order to diagnose COVID-19 \cite{zero}. In a section of the paper, there were some discussions about attention and transformer techniques. As it is mentioned, each of these methods has a lot of different techniques \cite{attendeep}. So, various attention and transformer models can diagnose COVID-19 in future works. 

\bibliographystyle{IEEEtran}
\bibliography{main}




\end{document}